\documentclass[lettersize,journal]{IEEEtran}
\usepackage{amsmath,amsfonts}
\usepackage{algorithmic}
\usepackage{algorithm}
\usepackage{array}
\usepackage{textcomp}
\usepackage{stfloats}
\usepackage{url}
\usepackage{verbatim}
\usepackage{graphicx}
\usepackage{cite}
\usepackage{subfigure} 
\usepackage{booktabs}
\usepackage{multirow}
\usepackage{mathrsfs}
\usepackage{amsmath}
\usepackage{amsfonts,amssymb} 

\begin{document}

\title{AMCEN: An Attention Masking-based \\ Contrastive Event Network for \\Two-stage Temporal Knowledge Graph Reasoning}

\author{Jing Yang, Xiao Wang, ~\IEEEmembership{Senior Member,~IEEE}, 
Yutong Wang, 
Jiawei Wang, 
Fei-Yue Wang, ~\IEEEmembership{Fellow,~IEEE}

% <-this % stops a space

\thanks{Jing Yang, Yutong Wang and Fei-Yue Wang are with the State Key Laboratory of Multimodal Artificial Intelligence Systems, Institute of Automation, Chinese Academy of Sciences, Beijing 100190, China and also with the School of Artificial Intelligence, University of Chinese Academy of Sciences, Beijing 100049, China (e-mail: yangjing2020@ia.ac.cn, yutong.wang@ia.ac.cn, feiyue.wang@ia.ac.cn).
}

\thanks{Xiao Wang is with School of Artificial Intelligence, Anhui University, Hefei 230031, China and also with Qingdao Academy of Intelligent Industries, Qingdao 266114, China. (E-mail: xiao.wang@ahu.edu.cn)}

\thanks{Jiawei Wang is with Zhejiang Lab, Hangzhou 311121, China (e-mail: wangjw@zhejianglab.com).}

\thanks{Manuscript received April 19, 2021; revised August 16, 2021. \\ \textit{(Corresponding author: Xiao Wang.)}}}

% The paper headers
\markboth{Journal of \LaTeX\ Class Files,~Vol.~14, No.~8, August~2021}%
{Shell \MakeLowercase{\textit{et al.}}: A Sample Article Using IEEEtran.cls for IEEE Journals}

% \IEEEpubid{0000--0000/00\$00.00~\copyright~2021 IEEE}
% Remember, if you use this you must call \IEEEpubidadjcol in the second
% column for its text to clear the IEEEpubid mark.

\maketitle

\begin{abstract}
Temporal knowledge graphs (TKGs) can effectively model the ever-evolving nature of real-world knowledge, and their completeness and enhancement can be achieved by reasoning new events from existing ones. However, reasoning accuracy is adversely impacted due to an imbalance between new and recurring events in the datasets. To achieve more accurate TKG reasoning, we propose an attention masking-based contrastive event network (AMCEN) with local-global temporal patterns for the two-stage prediction of future events. In the network, historical and non-historical attention mask vectors are designed to control the attention bias towards historical and non-historical entities, acting as the key to alleviating the imbalance. A local-global message-passing module is proposed to comprehensively consider and capture multi-hop structural dependencies and local-global temporal evolution for the in-depth exploration of latent impact factors of different event types. A contrastive event classifier is used to classify events more accurately by incorporating local-global temporal patterns into contrastive learning. Therefore, AMCEN refines the prediction scope with the results of the contrastive event classification, followed by utilizing attention masking-based decoders to finalize the specific outcomes. The results of our experiments on four benchmark datasets highlight the superiority of AMCEN. Especially, the considerable improvements in Hits@1 prove that AMCEN can make more precise predictions about future occurrences.
\end{abstract}


\section{Introduction}
\IEEEPARstart{K}{nowledge} in the real world undergoes continuous evolution over time, posing a challenge for static knowledge graphs (SKGs) to effectively describe the dynamic change. Therefore, temporal knowledge graphs (TKGs) are proposed to extend a time dimension on top of SKGs, contributing to depicting the constant evolution of temporal knowledge. TKGs consist of a sequence of event snapshots captured over different timestamps and present each fact as a quadruple \emph{(subject, relation, object, timestamp)}. However, knowledge graphs (KGs), whether TKGs or SKGs, are only capable of capturing partial and incomplete information from open systems \cite{ji2021survey}. To deal with it, knowledge graph reasoning (KGR) \cite{yang2021hackrl,yang2022hackgan,yang2022collective} has garnered considerable interest and refers to a process of inferring new facts from existing ones by mining underlying logical rules. The focus of this work is on the utilization of TKGs to forecast future events, which offers novel insights and perspectives for practical applications such as international negotiation \cite{DVN/28075_2015}, fault prediction \cite{su2020survey} and decision recommendation \cite{guo2020survey}.

\begin{figure}[!h]
    \centering  
    \includegraphics[width=1\linewidth]{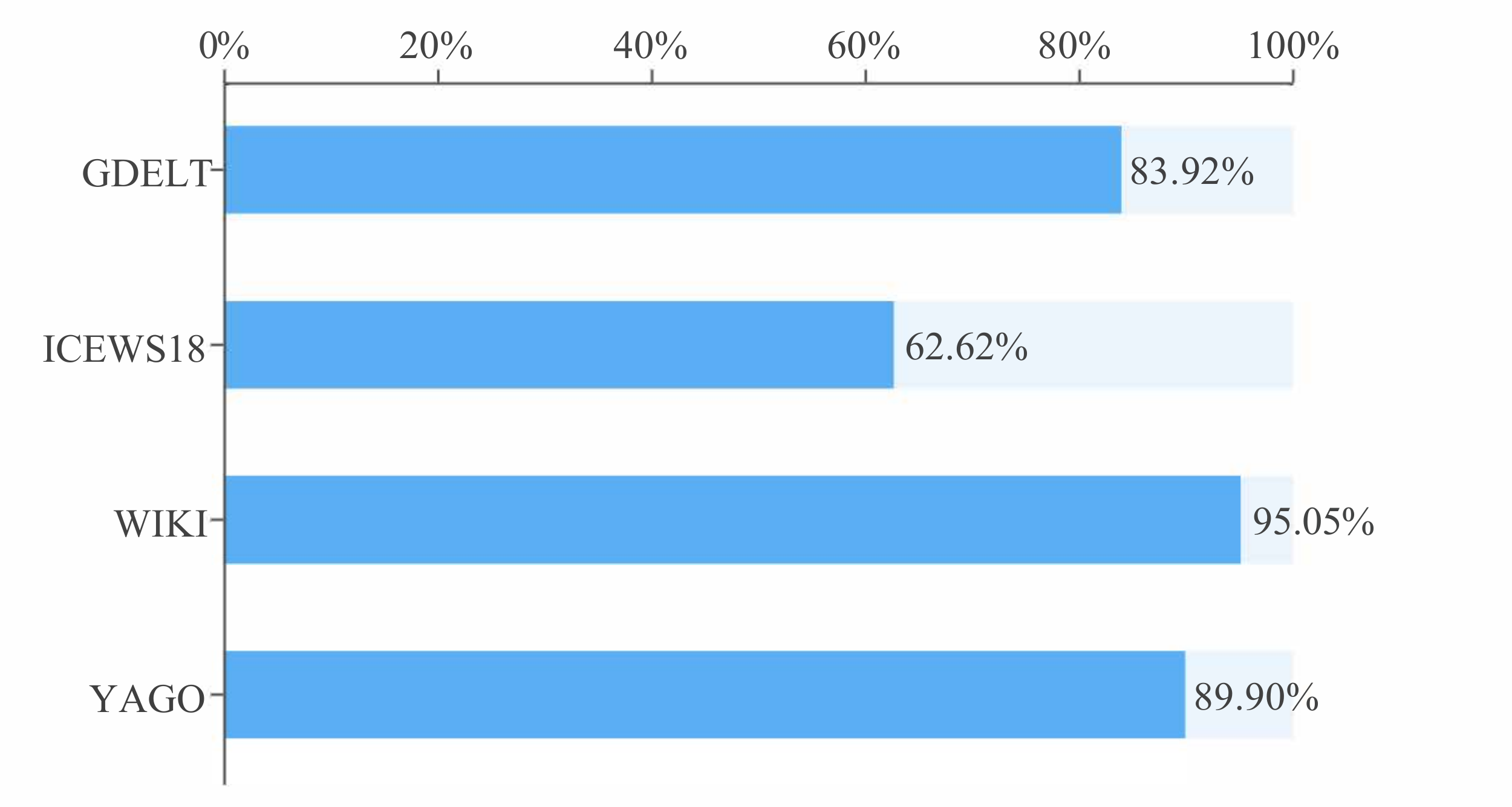}
    \caption{The proportion of recurring events in different datasets.} 
    \label{proportation1}
\end{figure}

Studies in \cite{jiang2016towards,dasgupta2018hyte, garcia2018learning} have expanded upon SKG embedding techniques \cite{bordes2013translating, wang2014knowledge, ji2015knowledge, nickel2011three, Yang2015DisMult, trouillon2016complex, dettmers2018convolutional, nguyen2017novel, schlichtkrull2018modeling,Nathani2019KBGAT,vashishth2019composition} by creating time-dependent scoring functions to evaluate the probability of derived facts. However, these methods fail to consider the multi-hop structural information within TKGs adequately. Methods in \cite{wu2020temp, jin2019recurrent,he2021hip, li2022tirgn, li2021temporal} combine graph neural networks (GNNs) and recurrent neural networks (RNNs) to encode neighbor features and sequential patterns of adjacent timestamps but lack consideration of historical repetitiveness. Other methods in \cite{zhu2021learning, xu2023cenet, liu2022net} incorporate the occurrence frequency of historical events on the global timeline. However, statistical data has demonstrated that the frequency of recurring events is generally higher than that of new events in the datasets, as depicted in Figure \ref{proportation1}. It signifies an imbalance between recurring and new events, which almost all methods have not considered and addressed. 
This causes these methods to disregard historical information that has no direct connection to the query but could significantly affect the likelihood of new events. To overcome this issue, we attempt to categorize the prediction of new and recurring events, thereby effectively distinguishing between the exploration of historical and non-historical information.

\begin{figure}[!h]
    \centering  
    \includegraphics[width=1\linewidth]{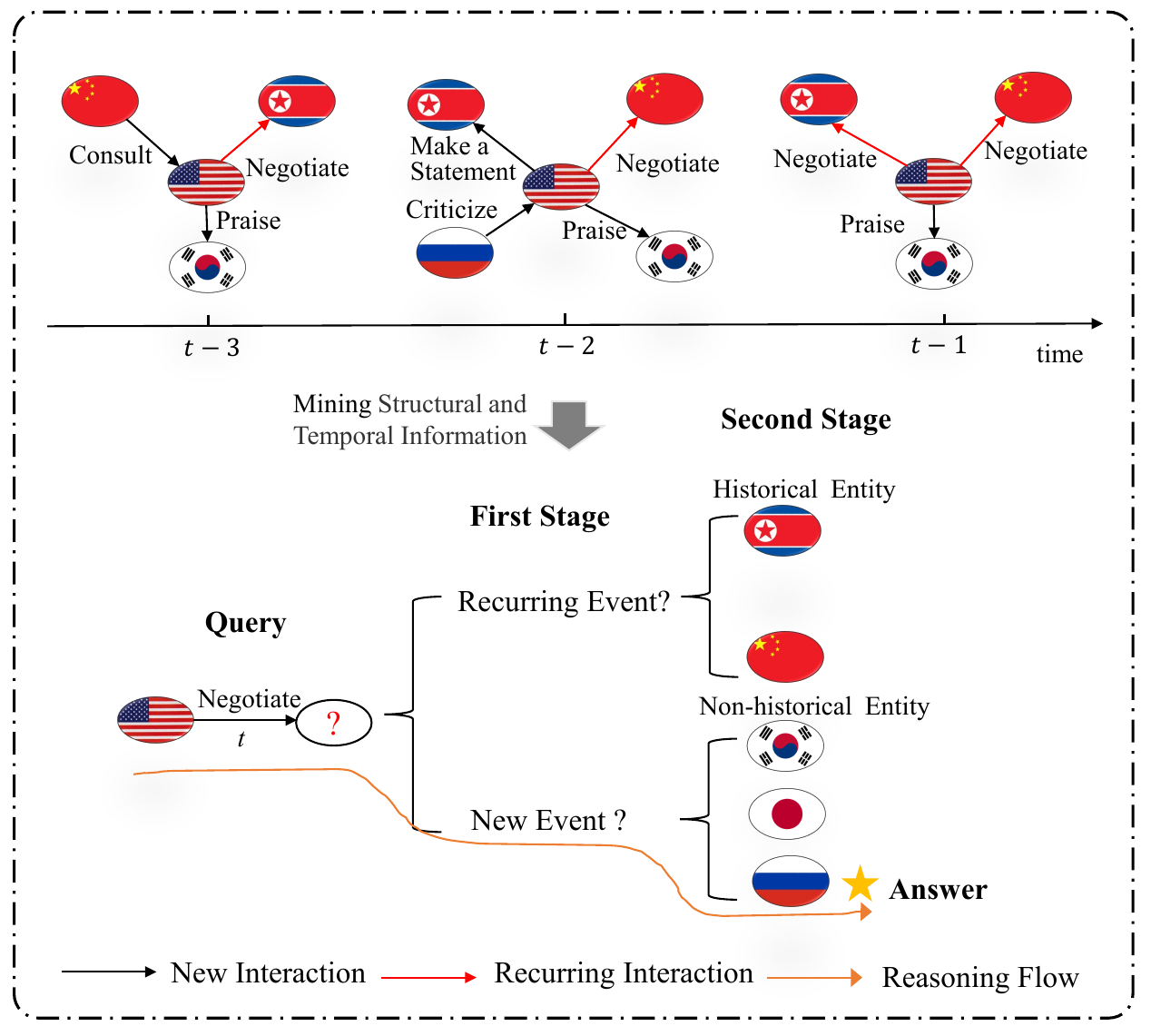}
    \caption{An illustration of a two-stage reasoning process.} 
    \label{illustration}
\end{figure}

Fig.\ref{illustration} illustrates the core idea of our proposed method, which is essentially a two-stage reasoning process involving initial classification followed by intra-class prediction. We take the example of predicting the missing objects of a query \emph{(the United States, Negotiate, ?, t)} in a future timestamp \emph{t}. Prior to the timestamp \emph{t}, \emph{the United States} has happened the \emph{“Negotiate”} relations with two countries, i.e., \emph{China} and \emph{North Korea}, which are known as historical entities. Conversely, other countries are non-historical entities. If, at the timestamp \emph{t}, \emph{the United States} still chooses to negotiate with one of these two countries, it is considered a recurring event. However, if \emph{the United States} opts to negotiate with a country other than \emph{China} and \emph{North Korea} at timestamp \emph{t}, it is deemed to be a new event. It is evident that current events are either recurring ones pertaining to historical entities, or new ones concerning non-historical entities. As recurring events occur frequently, existing models tend to misidentify new events as recurring events and select answers from historical entities, resulting in a decline in prediction accuracy.

\begin{figure*}[!h]
    \centering  
    \includegraphics[width=1\linewidth]{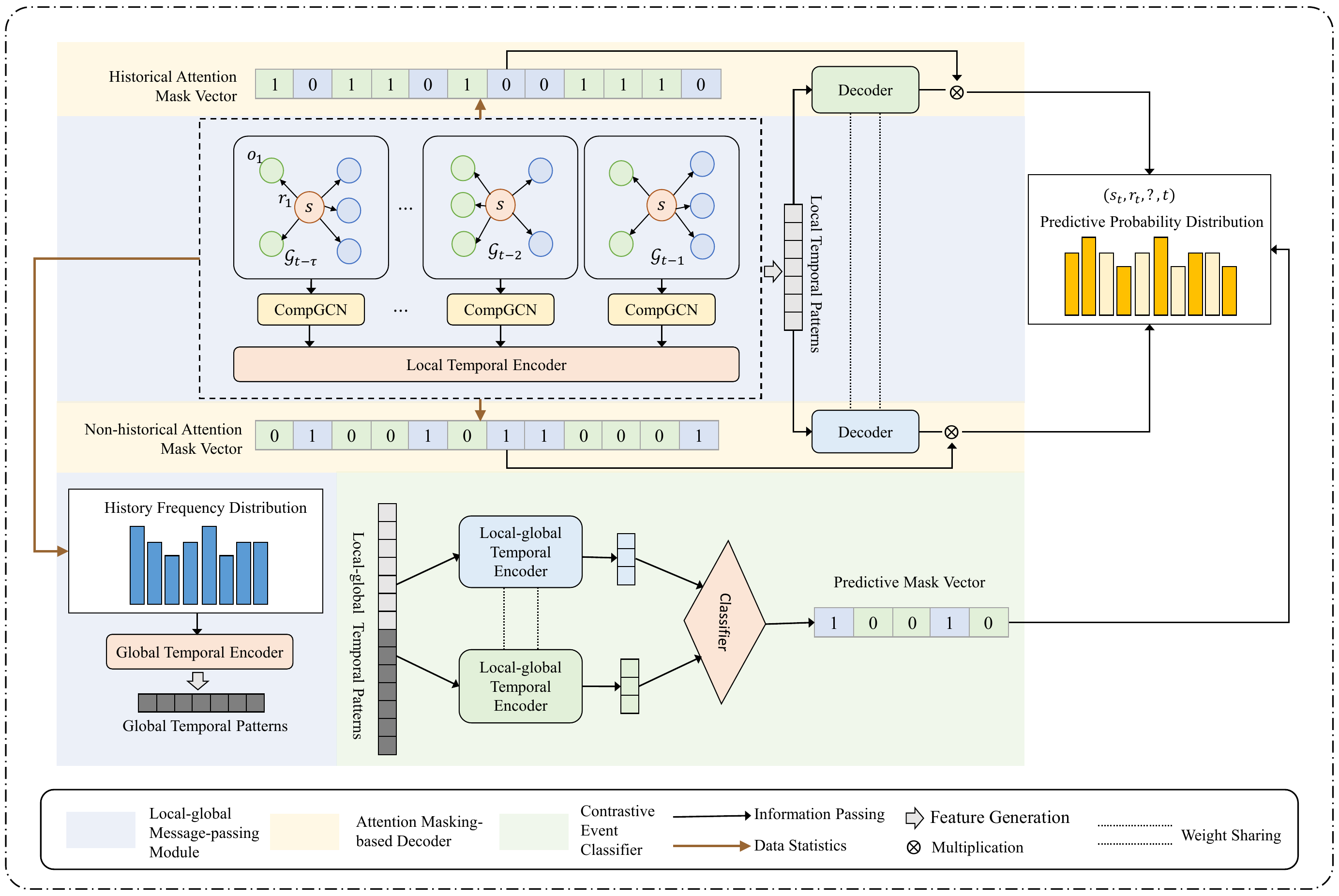}
    \caption{The overall architecture of AMCEN.} 
    \label{fig:model}
\end{figure*}

To deal with the imbalance issue and improve prediction accuracy, we subdivide the reasoning process into two steps from coarse to fine: 1) Determine whether the event to be predicted should be a recurring one or a new one; 2) Predict an object or a subject from the corresponding entity pool. As an example, we begin by classifying the query \emph{(the United States, Negotiate, ?, t)} as a new event. Subsequently, we utilize structural and temporal information to research the non-historical entity pool for the entity with the highest probability, which in this case is \emph{Russia}, serving as the answer to the query. This reasoning approach allows for a separation between the mining of historical and non-historical entities, which can improve the overall understanding of the underlying logic governing both new and recurring events.

Inspired by the above reasoning process, we propose a novel two-stage network for TKG reasoning. We design historical and non-historical attention mask vectors to separately learn latent factors leading to occurring and new events. To achieve in-depth mining of underlying logic, we use CompGCN \cite{vashishth2019composition} to update structural representations and employ self-attention \cite{vaswani2017attention} to iteratively capture local temporal features, which can ensure historical decay while avoiding the issue of feature stagnation caused by temporal sparsity. By incorporating local-global features into contrastive learning \cite{hassani2020contrastive}, we generate a predictive mask vector that helps to narrow down the prediction scope and thus improves prediction accuracy.

Our contributions are summarized as follows: 

\begin{itemize}
    \item We propose an attention masking-based contrastive event network (AMCEN) for temporal KGR, which comprehensively considers multi-hop structural, local temporal, and global temporal patterns based on CompGCN and self-attention while overcoming temporal sparsity and ensuring diminishing historical effects.
    \item We design historical and non-historical attention mask vectors to alleviate the issue of imbalanced distribution between new and recurring events in the datasets. The vectors separate the exploration of historical entities from that of non-historical entities, so it is possible to thoroughly explore all historical information, whether related to the query or unrelated.
    \item We conduct experiments on four datasets and results demonstrate that our method achieves superior or comparable to state-of-the-art performance on most metrics for entity prediction. Additionally, the results of ablation experiments confirm the effectiveness of the two-step reasoning.
\end{itemize}

The rest of this paper is organized as follows. In Section \uppercase\expandafter{\romannumeral2}, the related works are reviewed. Section  \uppercase\expandafter{\romannumeral3} introduces our task goal and related notations. Section \uppercase\expandafter{\romannumeral4} provides a summary of our model design and a detailed explanation of each individual component. In  Section \uppercase\expandafter{\romannumeral5}, we give some experimental results and discussions to verify the effectiveness of our model. Finally,  concluding remarks are drawn in Section \uppercase\expandafter{\romannumeral6}. 

\section{Related Works}
In this section, we first discuss the existing KGR works, consisting of static methods and temporal methods. Then several valuable techniques for improving KGR performance involved in our method are presented, including self-attention and masking as well as contrastive learning. 

\subsection{Static KGR Methods}
A considerable number of static KGR models have been developed to exploit the existing facts to infer queried facts. Path-based models \cite{lao2010relational,gardner2014incorporating,xiong2017deeppath} and rule-based models \cite{meilicke2018fine, galarraga2013amie,guo2016jointly} are explored and although they are more interpretable than other models, they suffer from limited expressive power and high spatiotemporal complexity. Therefore, embedding-based KGR has become the most prevalent approach, involving translation models \cite{bordes2013translating, wang2014knowledge, ji2015knowledge}, tensor decompositional models \cite{nickel2011three, Yang2015DisMult, trouillon2016complex} and neural network models \cite{dettmers2018convolutional, nguyen2017novel, schlichtkrull2018modeling,Nathani2019KBGAT,vashishth2019composition}. Their basic idea is to get a low-dimensional vector for each entity and relation, where the structural and semantic information of the original graph is maintained. TransE \cite{bordes2013translating}, the first translation model, regards relations between subjects and objects as translation vectors. As a representative tensor decompositional model, DisMult \cite{Yang2015DisMult} utilizes bi-linear diagonal matrices to model the symmetrical relations among latent factors. ComplEx \cite{trouillon2016complex} maps entities and relations into complex vector spaces, generalizing DisMult to antisymmetry relation modeling. Benefits from graph neural networks (GNNs)’ ability to represent graph structures with high quality, increasing works concentrate on GNN-based models, such as RGCN \cite{schlichtkrull2018modeling}, KBGAT \cite{Nathani2019KBGAT}, and CompGCN \cite{vashishth2019composition}. However, none of these models can capture the time-evolving properties of knowledge, leading to biased predictions of dynamic events.

\subsection{Temporal KGR Methods}
Dynamic KGR can be classified into interpolation and extrapolation based on whether the queried fact falls within the known time range. The former is also referred to as TKG completion. TTransE \cite{jiang2016towards}, an extension of TransE, encodes the temporal order information as a regularizer and temporal consistency information as constraints. HyTE \cite{dasgupta2018hyte} incorporates time as an explicit dimension in the entity-relation space by linking each timestamp to a corresponding hyperplane. TA-DistMult \cite{garcia2018learning} combines predicate sequence representation encoded by long short-term memory (LSTM) and DistMult as the scoring function. DySAT \cite{sankar2020dysat} uses structural and temporal self-attention layers to capture dynamic node representations. TeMP \cite{wu2020temp} integrates RGCN, a gated recurrent unit (GRU) and frequency-based gating techniques to overcome the temporal sparsity and variability while learning time-dependent representations. However, these methods are not specifically designed to predict future KGs and the utilization of them for extrapolation can lead to larger reasoning errors.

The latter is the focus of this paper, aiming at predicting future events and many related models have been yielded and discussed in the following. Due to their ability to mine temporal trends, it is natural to apply recurrent neural networks (RNNs) and their variants in sequential inference tasks. Re-Net \cite{jin2019recurrent} formulates the occurrence of an event as a probability distribution by conditioning on preceding temporal sequences of events and employs an RNN and a neighborhood aggregator to encode temporal and structural dependency. HIP network \cite{he2021hip} leverages a GRU to obtain the relation transition representation. TiRGN \cite{li2022tirgn} uses a GRU as a local recurrent graph encoder and designs a global history encoder for learning history repetitive patterns. RE-GCN \cite{li2021temporal} models all the historical information into evolutional representations based on GRU and relation-aware GCN. TITer\cite{sun2021timetraveler} is an explainable  temporal-path-based reinforcement learning model that defines a relative time encoding function to represent the time information and utilizes LSTM to encode historical paths for agents. In addition to the above RNN-based models, there are also some models that introduce time information without an RNN framework. GyGNet \cite{zhu2021learning} introduces a time-aware copy-generation mechanism to select repeated facts from the historical vocabulary for extrapolated graph reasoning. xERTE \cite{han2021explainable} is an interpretable reasoning framework to extract query-related subgraphs by iteratively sampling temporal neighbors. CEN \cite{li2022complex} employs curriculum learning to learn evolutional patterns with the characteristics of length diversity by utilizing a length-aware convolution neural network (CNN). The design of DA-Net \cite{liu2022net} draws inspiration from human decision-making mechanisms, and it leverages distributed attention to model the dynamic distributions of frequently occurring historical facts in a dual process. However, due to the unbalanced ratio of recurring and new events, these models tend to prioritize mining query-related historical information and ignore the query-unrelated parts, despite the latter's vital potential impact on the occurrence of new events. CENET \cite{xu2023cenet} proposes historical and non-historical dependency to  overcome biases towards historical entities. But this method is insufficient in mining historical information due to the lack of consideration for structural information and sequential patterns of facts. 

\subsection{Self-attention and Masking}

In the field of deep learning, attention is introduced by Google DeepMind in 2014 \cite{mnih2014recurrent}, which refers to the selective focus and processing of information with different importance. Self-attention is a commonly used attention mechanism that calculates the internal relevance weight of each input element by interacting with one another, resulting in improved representations. Transformer \cite{vaswani2017attention}, which relies entirely on self-attention, utilizes a multi-head attention mechanism to simultaneously attend to information from various representation subspaces at different positions. Additionally, it modifies the self-attention sublayer in the decoder by masking to prevent any given position from attention to subsequent positions. Pretrained models like BERT design a masked language modeling task, which is capable of effectively extracting valuable information from the context by randomly masking one or more words in the text \cite{kenton2019bert}. Recently, self-attention has been applied to KGR, such as KBGAT \cite{Nathani2019KBGAT}, DySAT \cite{sankar2020dysat}, TeMP \cite{wu2020temp} , and HIP network \cite{he2021hip}, which selectively capture neighbor features or local temporal features. TeMP \cite{wu2020temp} proves self-attention can be regarded as an imputation approach that integrates stale representations with temporal representations for inactive entities. Obviously, applying self-attention to TKGs can not only more easily capture important features in the KG sequence, but also avoid the problem of feature obsolescence caused by temporal sparsity. Masking allows for deliberate control over the model's attention towards specific information, effectively blocking the model from accessing unwanted information.

\subsection{Contrastive Learning }

Contrastive learning \cite{hassani2020contrastive} is a discriminative paradigm in self-supervised learning \cite{jaiswal2020survey} that allows the utilization of self-defined pseudo labels as supervision to learn representations that are useful for various downstream tasks. It aims to encourage similar samples to cluster together while pushing diverse samples apart from each other. It has achieved great success in various downstream tasks such as image classification, object detection and future prediction \cite{chen2020simple,liu2022deep,gutmann2012noise}. Recent works have applied the learning mechanism to inductive reasoning for static KG. RPC-IR \cite{pan2021learning} is an interpretable inductive reasoning model that constructs positive and negative relational paths between two entities for a contrastive strategy. SNRI \cite{xu2022subgraph} maximizes the mutual information (MI) of subgraph and graph representations to globally capture neighboring relations. CENET \cite{xu2023cenet} introduces contrastive learning for temporal reasoning to differentiate whether the query relies more on historical or non-historical events. However, the query encoder in this model does not utilize fine-grained temporal information, resulting in poor presentation performance and thus misleading the subsequent classification. Therefore, our method introduces local-global temporal patterns into the contrastive learning framework, which can enhance the performance of subsequent classifiers. 

% \section{Method}
% In this section, we start by introducing our task goal and related notations. Then we provide a summary of our model design and a detailed explanation of each individual component. 

\section{Notations and Task Definition}
In a TKG, let $\mathcal E$, $\mathcal R$, $\mathcal T$ and $\mathcal F$ be a finite set of entities, relations, timestamps and events, respectively. A TKG can be described as a series of SKGs that occur at different timestamps $\mathcal G= \lbrace{\mathcal G_1,\mathcal G_2,...,\mathcal G_t,...,\mathcal G_T }\rbrace$, where $\mathcal G_t=\lbrace{\mathcal E,\mathcal R,\mathcal F_t}\rbrace$ denotes the SKG at the timestamp $t\in \mathcal T$ and $T$ is the size of $ \mathcal T$. $\mathcal F_t$ contains all the events that happen at the timestamp $t$. A event can be represented in the form of a quadruple $(s,r,o,t)$ that means a subject $s\in \mathcal{E}$ has a relation $r\in \mathcal{R}$ with an object $o\in \mathcal{E}$ at the timestamp $t$. The bolded letters $\mathbf{s},\mathbf{r},\mathbf{o},\mathbf{t} \in \Bbb{R}^d$ are used to stand for $d$-dimensional embedding vectors of $s,r,o,t$, respectively.

An event $(s,r,o,t)$ is a recurring event if $\exists (s,r,o,k) \in \mathcal G_k$ and $k<t$. Otherwise, it is a new event. Correspondingly, we define the historical entities of a query $q=(s,r,?,t)$ as $H_t^{s,r}=\lbrace{o|(s,r,o,k)\in \mathcal G_k} \rbrace$ and its non-historical entities as $N_t^{s,r}=\lbrace{o| o\in \mathcal E \  \mathtt{and} \ o\notin H_t^{s,r} } \rbrace$. Our task is to complete the missing part for an object query $(s,r,?,t)$ or a subject query $(?,r,o,t)$ by using historical information before timestamp $t$, i.e., $G_{1:t-1}$. Without loss of generality, we elaborate on our temporal KGR approach AMCEN using an object query as an example. 

\section{model architecture}
% The overall architecture of AMCEN is shown in Fig. \ref{fig:model}, consisting of four main components: structural encoder, temporal encoder, historical/nonhistorical attention mask and decoder, as well as contrastive learning and classifier. Of these, the first two can be collectively referred to as a message-passing module. Specifically, the structural encoder integrates the interactions of co-occurring events to learn multi-hop embeddings of partial subgraphs by employing a CompGCN-based aggregator;  the temporal encoder is composed of an attention-based local temporal encoder and a frequency-based global temporal encoder to capture the sequential and repetitive patterns of temporal evolution. Decoder scores each candidate entity by computing similarity with queries to decode the future graphs of events, where historical and non-historical attention masks are designed to control the distribution of its attention. Contrastive learning combines local and global time-dependent features to prepare for subsequent better classification of queries that aims at narrowing down the predictive scope. On the basis of this structure, we perform two-stage temporal knowledge graph reasoning to generate predicted graphs for future timestamps. 

The overall architecture of AMCEN is shown in Fig. \ref{fig:model}, consisting of three main components: a local-global message-passing module, two attention masking-based decoders, and a contrastive event classifier. The local-global message-passing module is composed of structural encoders and temporal encoders. Structural encoders integrate the interactions of co-occurring events to learn multi-hop embeddings of partial subgraphs by employing a CompGCN-based aggregator. Temporal encoders involve an attention-based local temporal encoder and a frequency-based global temporal encoder to capture the sequential and repetitive patterns of temporal evolution. Attention masking-based decoders score each candidate entity by computing similarity with queries to decode the future graphs of events, where historical and non-historical attention mask vectors are designed to control the distribution of their attention. The contrastive event classifier combines local and global time-dependent features to obtain contrastive representations, and thus feed them to a binary classifier for better classification of queries. Based on this structure, we perform two-stage temporal knowledge graph reasoning to generate predicted graphs for future timestamps.

\subsection{Local-global Message-Passing Module}

\subsubsection{Structural Encoder}
To better convey information between co-occurring events through shared entities, we leverage multi-relation CompGCN \cite{vashishth2019composition} to aggregate multi-hop neighbor features and thus jointly embed both nodes and relations in an SKG. In addition, this approach can also alleviate over-parameterization with an increasing number of relations by using basis decomposition and sharing relation embeddings across layers, which are the main problem that traditional RGCNs suffer from. The CompGCN update equation is presented as:

\begin{equation}\label{eq1}
\mathbf{h}_{s,t}^{l+1}=f_{stru}(\sum_{(r,o)\in \mathcal N(s)}(\mathbf{W}^l_{\lambda(r)} \phi(\mathbf{h}_{o,t}^{l},\mathbf{h}_{r,t}^{l} )))
\end{equation}

\begin{equation}\label{eq2}
\mathbf{h}_{r,t}^{l+1}=\mathbf{W}^l_{rel}\mathbf{h}_{r,t}^{l}
\end{equation}

\begin{equation}\label{eq3}
\mathbf{W}_{\lambda(r)}=\left\{
\begin{array}{rcl}
\mathbf{W}_O && r \in \mathcal R\\
\mathbf{W}_I && r \in \mathcal R_{inv}\\
\mathbf{W}_S && r \in \mathcal S 
\end{array} \right.
\end{equation}

\noindent where $\mathcal N(s)$ is a set of immediate neighbors of $s$ for its outgoing edges $r$ to nodes $o$; $\mathbf{{h}}_{s,t}^{l}$, $\mathbf{{h}}_{o,t}^{l}$ and $\mathbf{{h}}_{r,t}^{l}$ stand for the $l^th$-layer embeddings of entities $s,o$ and relations $r$ at timestamp $t$, and $\mathbf{{h}}_{s,t}^{0}=\mathbf{s}$, $\mathbf{{h}}_{o,t}^{0}=\mathbf{o}$ and $\mathbf{{h}}_{r,t}^{0}=\mathbf{r}$ are initial embeddings; $\mathbf{W}_{\lambda(r)} \in \Bbb{R}^{d\times d}$ and $\mathbf{W}^l_{rel} \in \Bbb{R}^{d\times d}$ denote learnable transformation matrix and $\mathbf{W}_{\lambda(r)}$ is relation-type-specific, involving $\mathbf{W}_{O}$, $\mathbf{W}_{I}$ and $\mathbf{W}_{S}$; $f_{stru}(x)=softmax(ReLU(x))$ is the activation function; $\mathcal R_{inv}=\lbrace{r^{-1}|r\in \mathcal R}\rbrace$
indicates the inverse relations and  $\mathcal S$ represents the self loop; $\phi:\Bbb{R}^{d} \times \Bbb{R}^{d} \to \Bbb{R}^{d}$ is a composition operator, including subtraction \cite{bordes2013translating}, multiplication \cite{yang2014embedding}, circular-correlation \cite{nickel2016holographic}, ConvE \cite{dettmers2018convolutional}, etc. After L-layer aggregation, the outputs of CompGCN are $\mathbf{z}_{s,t}=\mathbf{{h}}_{s,t}^{L}$ and $\mathbf{z}_{r,t}=\mathbf{{h}}_{r,t}^{L}$, which capture L-hop structural dependencies. 

\subsubsection{Temporal Encoder}
The evolution of snapshots is sequential in adjacent timestamps, and some events are repeated along the global timeline. The temporal encoder is dedicated to learning time-aware representation from both local and global perspectives. Specifically, We count the frequencies $\mathbf{F}_{s,r}^{t} \in \Bbb{R}^{|\mathcal E|}$ of different objects that have ever occurred for the query $(s,r,?,t)$. And then we encode them as global features $\mathbf{H}_{s,r}^{global_t}$ to portray the recurrence pattern of events:

\begin{equation}\label{eq4}
\mathbf{F}_{s,r}^{t}(o)=\sum_{k<t}|\lbrace{o|(s,r,o,k)\in \mathcal G_k} \rbrace|
\end{equation}

\begin{equation}\label{eq5}
\mathbf{H}_{s,r}^{global_t}=f_F(\mathbf{W}_F\mathbf{F}_{s,r}^{t}+\mathbf{b}_F)
\end{equation}

\noindent where ${W}_F \in \Bbb{R}^{b\times|\mathcal E|}$, $\mathbf{b}_F \in \Bbb{R}^b$ are learnable weights and bias, $f_F(x)$ is a tanh activation function. 

In terms of local patterns, we use self-attention to selectively integrate the sequence of historical information within the time windows based on the snapshot at $t-1$ in an iterative manner (see Equations (\ref{eq6}-\ref{eq7})), that is, attentive pooling. This is because the future snapshot at timestamp $t$ must evolve from the snapshot at $t-1$. In other words, the likelihood of similarity between the two snapshots is relatively high. Therefore, the snapshot at $t-1$ is of great  importance to predict the future snapshot at $t$. It is obvious that the iterative approach results in the decreasing proportion of historical information as the time distance increases, which is consistent with diminishing historical effects. Meanwhile, the attentive pooling can avoid stale representation sharing across multiple time steps because of the long-term inactivity of entities, namely, time sparsity \cite{wu2020temp}. This reason is that it always conjoins the previously active state of entities. 

\begin{equation}\label{eq6}
\begin{split}
\mathbf{attention}_{s,j}=softmax(\frac{(\mathbf{z}_{s,t-1}\mathbf{W}_q)(\mathbf{x}_{s,t-j}\mathbf{W}_k)^T}{\sqrt{d}}) \\
j=2,3,\dots,\tau
\end{split}   
\end{equation}

\begin{equation}\label{eq7}
\mathbf{x}_{s,{(t-1)}^-}=FFN(\sum_{j=\tau-1}^2 \mathbf{attention}_{s,j}(\mathbf{x}_{s,t-j}\mathbf{W}_v)) 
\end{equation}

\noindent where $\mathbf{W}_q$, $\mathbf{W}_k$ and $\mathbf{W}_v \in \Bbb{R}^{d \times \frac{d}{m}}$ are learnable linear projections, $FFN(\cdot)$ represents a feed-forward network with $d$ hidden units, ${(t-1)}^-\in \lbrace{t-\tau,\dots,t-2,t-1 }\rbrace$ and $\tau$ denotes the size of the time window. $m$ is the number of attention heads. We set multiple attention heads ($h$ stands for the number of them) to incorporate the importance of adjacent snapshots from multiple perspectives. 

To comprehensively consider various information in the previous moments, we merge structural features $\mathbf{z}_{s,t-1}$ at timestamp $t-1$ with time-aware representations $\mathbf{x}_{s,{(t-1)}^-}$ before $t-1$ using a parameter $\beta$. Obviously, this approach is necessary due to the significance of the snapshot at $t-1$. Finally, the time-dependent representation $\mathbf{x}_{s,t}$ of a subject is defined as: 

\begin{equation}\label{eq8}
\mathbf{x}_{s,t}=\beta \ast \mathbf{z}_{s,t-1}+ (1-\beta) \ast \mathbf{x}_{s,{(t-1)}^-}
\end{equation}

Similarly, we adapt Equations (\ref{eq4}-\ref{eq8}) to the calculation of time-dependent relation representations $\mathbf{x}_{r,t}$. Finally, the local temporal patterns of the query $(s,r,?,t)$ can be given as: 

\begin{equation}\label{eq9}
\mathbf{H}_{s,r}^{local_t}= \mathbf{x}_{s,t} \oplus \mathbf{x}_{r,t} \oplus \mathbf{t}
\end{equation}

\noindent where $\oplus$ is a concatenation operator. 

\subsection{Attention Masking-based Decoder}
Actually, the queried event is either a recurring one or a new one. The prevalence of recurring events over new events often fosters a tendency to predict outcomes based on historical entities. However, it is indispensable to predict the occurrence of new events, which is of great importance in reality. That is, non-historical entities corresponding to new events should not be ignored. To mitigate the uneven distribution of event types, we design historical and non-historical attention mask vectors to separate the exploration of historical entities from that of non-historical entities, due to the agnostic event type of the query in advance. A historical attention mask vector $\mathbf{M}_{s,r}^{his_t} \in \Bbb{R}^{|\mathcal{E}|}$ and a non-historical attention mask vector $\mathbf{M}_{s,r}^{nhis_t} \in \Bbb{R}^{|\mathcal{E}|}$ are defined as: 

\begin{equation}\label{eq10}
\mathbf{M}_{s,r}^{his_t}(o)=\bf I \mathit {(F_{s,r}^{t}(o)>0)}
\end{equation}
\begin{equation}\label{eq11}
\mathbf{M}_{s,r}^{nhis_t}(o)=\bf I \mathit {(F_{s,r}^{t}(o)==0)}
\end{equation}

\noindent where $\bf{I} \mathit{(x)}$ is an indicator function that return 1 if $\mathit{x}$ is true and 0 otherwise.

To predict future KGs, we project the query and candidate entities onto a unified space, which allows us to implement a decoder to compute their similarity. The decoding function $\mathbf{D}_{s,r}^t \in \Bbb{R}^{|\mathcal{E}|}$ is represented as follows:

\begin{equation}\label{eq12}
\mathbf{D}_{s,r}^t=softmax(\frac{((\mathbf{x}_{s,t} \oplus \mathbf{x}_{r,t})\mathbf{W}_q^D)(\mathbf{x}_{o,t}\mathbf{W}_k^D)^T}{\sqrt{d}}) 
\end{equation}

\noindent where $\mathbf{W}_q^D \in \Bbb{R}^{d \times 2d}$ and $\mathbf{W}_k^D \in \Bbb{R}^{d \times d}$ are learnable linear projection matrices according to the query and keys. 

For the exploration of factors influencing new events, we should focus on investigating non-historical entity pools by masking historical entities. Conversely, for recurrent events, our attention should shift to exploring historical entity pools while excluding non-historical entities. The selective attention functions ($\mathbf{C}_{s,r}^{his_t}$, $\mathbf{C}_{s,r}^{nhis_t} \in \Bbb{R}^{|\mathcal{E}|}$) are defined as: 

\begin{equation}\label{eq13}
\mathbf{C}_{s,r}^{his_t}= \mathbf{M}_{s,r}^{his_t} \cdot \mathbf{D}_{s,r}^t
\end{equation}
\begin{equation}\label{eq14}
\mathbf{C}_{s,r}^{nhis_t}= \mathbf{M}_{s,r}^{nhis_t} \cdot \mathbf{D}_{s,r}^t
\end{equation}

It is worth noting that $\mathbf{C}_{s,r}^{his_t}$ and $\mathbf{C}_{s,r}^{nhis_t}$ share learnable parameters, which contributes to the optimization of different parts of the same network through selective attention. The training objective is to minimize the sum of two cross-entropy losses $\mathcal{L}_{mc}$: 

\begin{equation}\label{eq15}
\begin{split}
\mathcal{L}_{mc} &= CrossEntropyLoss(\mathbf{C}_{s,r}^{his_t}, o_{gt})\\
&+CrossEntropyLoss(\mathbf{C}_{s,r}^{nhis_t}, o_{gt})    
\end{split}
\end{equation}

\noindent where $o_{gt}$ stands for the ground truth of the query $(s,r,?,t)$. The predicted probabilities $\mathbf{P}_{s,r}^t \in \Bbb{R}^{|\mathcal{E}|}$ for each object are computed according to the following equation. 

\begin{equation}\label{eq16}
\mathbf{P}_{s,r}^t(o)=\frac{1}{2}({C}_{s,r}^{his_t}(o)+{C}_{s,r}^{nhis_t}(o))     
\end{equation}

\begin{table*}[t]
  \centering
  \caption{Statistics of Four Datasets. }
  \label{tab:datasets}
  \setlength{\tabcolsep}{0.5mm}{
  \resizebox{1\linewidth}{!}{
  \begin{tabular}{c|c|c|c|c|c|c|c|c|c|c|c}
    \toprule
    \multirow{2}{*}{Datasets} & \multirow{2}{*}{Entities} & \multirow{2}{*}{Relations} & \multirow{2}{*}{Granularity} & Time & \multicolumn{3}{c|}{Data Splitting} & \multicolumn{4}{c}{Proportion of New Events in} \\
    &&&&Granules&Training&Validation&Test&
  Training& Validation& Test& All \\
    \midrule
    YAGO & 10623 & 10 & 1 year & 189 & 161,540 & 19,523 & 20,026
            &  10.31\% & 11.31\% & 7.21\% & 10.10\%\\
    WIKI & 12,554 & 24 & 1 year & 232 & 539,286 & 67,538 & 63,110 & 2.81\% & 14.61\% & 12.88\% & 4.95\%\\
    ICEWS18 & 23,033 & 256 & 24 hours & 304 & 373,018 & 45,995 
            & 49,545 & 39.39\% & 29.37\% & 29.68\% & 37.38\%\\
    GDELT & 7,691 & 240 & 15 mins & 2,751 & 1,734,399 & 238,765 
            & 305,241 & 17.81\% & 10.82\% & 10.39\% & 16.08\%\\
    \bottomrule
  \end{tabular}
}}

\end{table*}

% \begin{table*}[t]
%   \centering
%   \caption{Statistics of Four Datasets. }
%   \label{tab:datasets}
%   \setlength{\tabcolsep}{0.5mm}{
%   \resizebox{0.85\linewidth}{!}{
%   \begin{tabular}{c|c|c|c|c|c|c|c|c|c|c}
%     \toprule
%     \multirow{2}{*}{Datasets} & \multirow{2}{*}{Entities} & \multirow{2}{*}{Relations} & \multirow{2}{*}{Granularity} & Time & \multicolumn{3}{c|}{Data Splitting} & Proportion of New Events \\
%     &&&&Granules&Training&Validation&Test& in Test Set \\
%     \midrule
%     YAGO & 10623 & 10 & 1 year & 189 & 161,540 & 19,523 & 20,026
%             &  7.21 \%\\
%     WIKI & 12,554 & 24 & 1 year & 232 & 539,286 & 67,538 & 63,110          & 12.88\%\\
%     ICEWS18 & 23,033 & 256 & 24 hours & 304 & 373,018 & 45,995 
%             & 49,545 & 29.68\%\\
%     GDELT & 7,691 & 240 & 15 mins & 2,751 & 1,734,399 & 238,765 
%             & 305,241 & 10.39\% \\
%     \bottomrule
%   \end{tabular}
%   }}

% \end{table*}

\subsection{Contrastive Event Classifier }
In this section, our goal is to predict the event type of query to narrow down the scope of candidate entities. To improve classification accuracy, we combine local-global patterns with contrastive learning to generate query contrastive representations $\mathbf{v}_q \in \Bbb{R}^d $. These representations have the characteristic of small intra-class distances and large inter-class distances, which can better assist subsequent classification tasks. First, we label each event sample $q_i$ in the dataset $\mathcal{M}=\lbrace{q_i}\rbrace, i=1,2,3,\dots,n$, where $n$ represents batch size: $I_q$ is set to $1$ if a sample is a recurring event, and $0$ otherwise. Therefore, for any given query $q_i$, we can obtain a set of similar queries that belong to the same category $N(q_i)=\lbrace{q_j|I_{q_j}=I_{q_i}}\rbrace$. Then, we optimize the contrastive learning model by minimizing the contrastive loss $\mathcal{L}_{bc}$ as follows: 

\begin{equation}\label{eq17}
\mathbf{v}_q=FFN(\mathbf{H}_{s,r}^{local_t} \oplus \mathbf{H}_{s,r}^{global_t})  
\end{equation}

\begin{small}
\begin{equation}\label{eq18}
\begin{split}
\mathcal{L}_{bc} &=\sum_{q_i \in \mathcal{M}}{\frac{-1}{|N(q_i)|}}\sum_{q_j \in N(q_i)}log(\frac{exp(\mathbf{v}_{q_i}\cdot \mathbf{v}_{q_j}/\mu)}{\sum_{q_k\in \mathcal{M} \setminus{\lbrace q_i \rbrace}}exp(\mathbf{v}_{q_k}\cdot \mathbf{v}_{q_i} /\mu)})  
\end{split} 
\end{equation}
\end{small}

\noindent where $ \mathcal{M}\setminus{\lbrace q_i \rbrace}$ denotes the removal of the $q_i$ element from the set $\mathcal{M}$.

Subsequently, we feed the optimized contrastive representations $\mathbf{v}_q$ to a binary classifier with a cross-entropy loss. It is worth noting that before training the classifier, we must freeze the parameters of the other structures, including the encoders, decoder and contrast learning model. The output of the classifier is a predicted label as $\hat{I_q}$ that indicates the category of event that the query pertains to:

\begin{equation}\label{eq19}
\hat{I_q}=Classifier(\mathbf{v}_q)  
\end{equation}

Based on $\hat{I_q}$, a corresponding predictive mask vector $\mathbf{M}_q^{pred} \in \Bbb{R}^{|\mathcal{E}|}$ for the given query is generated as:

\begin{equation}\label{eq20}
\mathbf{M}_q^{pred}(o)=\mathbf{I}{({\hat{I}_q}>\frac{1}{2})} \cdot \mathbf{I} {(F_{s,r}^{t}(o)>0)}
\end{equation}

Equation (\ref{eq20}) indicates that if the given query is predicted as a recurring event ($\hat{I}_q>\frac{1}{2}$), the corresponding positions of historical entities are set to $1$ and vice versa.

\subsection{Inference and Training Strategy}
The predicted probability distribution is refined by applying predictive mask vectors to obscure any entities that might introduce interference:

\begin{equation}\label{eq21}
\mathbf{P}(o|s,r,t,\mathcal{G}_{1:t-1})= \mathbf{P}_{s,r}^t(o) \cdot \mathbf{M}_q^{pred}(o)
\end{equation}

The object with the highest probability is chosen as the final prediction of the query $(s,r,?,t)$:

\begin{equation}\label{eq22}
\hat{o}=argmax_{o \in \mathcal{E}} \mathbf{P}(o|s,r,t,\mathcal{G}_{1:t-1})
\end{equation}

To this end, the entire training process of AMCEN is composed of two stages. The first stage is oriented towards multi-classification tasks, where the local-global message-passing module, attention masking-based decoders and the contrastive learning part of contrastive event classifiers are trained by minimizing the following total loss:

\begin{equation}\label{eq23}
\mathcal{L}=\lambda \ast \mathcal{L}_{mc}+(1-\lambda) \ast \mathcal{L}_{bc}
\end{equation}

\noindent where $\lambda$ is a trade-off hyper-parameter between decoders and contrastive learning.  

After fixing the trained parameters of the first stage, we optimize a binary classifier with a binary cross-entropy loss and thus obtain predictive mask vectors. However, with respect to validation and test, the above two stages should be flipped: the first stage is to determine whether answers should be chosen from historical or non-historical entities through the local-global message-passing module and contrastive event classifier; the second stage is to obtain the final results from the corresponding entity pool through the local-global message-passing module and attention masking-based decoders. 

\begin{figure}[!h]
    \centering  
    \includegraphics[width=1\linewidth]{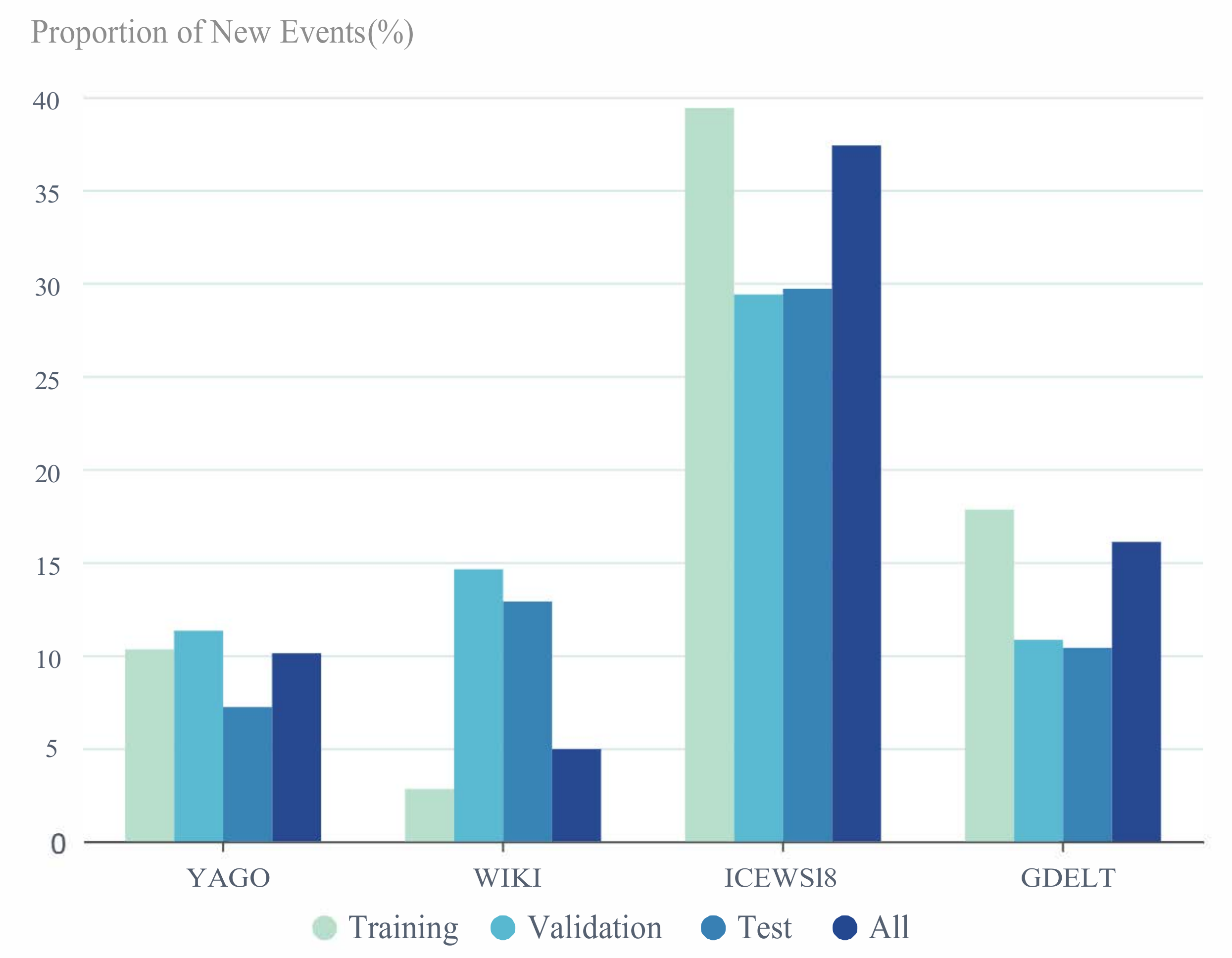}
    \caption{The distribution of new events in different datasets.} 
    \label{fig:proportion}
\end{figure}

\section{Experiments}

\subsection{Datasets and Baselines}

\subsubsection{datasets} We evaluate the performance of AMCEN for entity prediction on four benchmark datasets, consisting of ICEWS18\cite{DVN/28075_2015}, GDELT\cite{leetaru2013gdelt}, WIKI\cite{leblay2018deriving} and YAGO\cite{mahdisoltani2014yago3}. ICEWS18 and GDELT are event-based TKGs from the Integrated Crisis Early Warning System as well as the Global Database of Events, language, and Tone, respectively, where events are collected and updated on a relatively short time granularity. YAGO and WIKI are two fact-based knowledge bases that contain time information, where each fact can endure over an extended period of time. Following previous work\cite{han2021explainable,he2021hip,li2022tirgn, jin2019recurrent,li2022complex,li2021temporal, sun2021timetraveler, zhu2021learning, liu2022net,xu2023cenet}, we chronologically split each dataset into training, validation and test sets in proportions of 80\%, 10\% and 10\%. According to our statistics (see Fig. \ref{fig:proportion}), it is found that ICEWS18 and GDELT contain a higher proportion of new events compared to YAGO and WIKI on the whole datasets. ICEWS18 has the most relationship and entity types. On WIKI, the proportion of new events in the training set is much smaller than that in the validation and testing sets. More details about the four datasets are shown in Table \ref{tab:datasets}.

\subsubsection{Baselines}
Our proposed AMCEN method is compared with a broad spectrum of existing approaches, including static KGR methods and temporal KGR methods. TransE \cite{bordes2013translating}, DisMult \cite{Yang2015DisMult}, ComplEx \cite{trouillon2016complex}, ConvE \cite{dettmers2018convolutional}, RotatE \cite{sun2019rotate}, RGCN \cite{schlichtkrull2018modeling} and CompGCN \cite{vashishth2019composition} are selected as static reasoning baselines. TTransE \cite{jiang2016towards}, HyTE \cite{dasgupta2018hyte},  TA-DistMult \cite{garcia2018learning} , DySAT \cite{sankar2020dysat} and TeMP \cite{wu2020temp} are temporal KGR methods designed for TGR completion. Extrapolation-targeted temporal reasoning baselines consist of RNN-based models and RNN-agnostic models. The former involves Re-Net \cite{jin2019recurrent}, HIP network \cite{he2021hip}, TiRGN \cite{li2022tirgn}, RE-GCN \cite{li2021temporal} and TITer\cite{sun2021timetraveler}. The latter includes GyGNet \cite{zhu2021learning}, xERTE \cite{han2021explainable}, CEN \cite{li2022complex}, DA-Net \cite{liu2022net} and CENET \cite{xu2023cenet}. The results of other baselines are adopted from \cite{liu2022net}, except TiRGN \cite{li2022tirgn} and CENET \cite{xu2023cenet} report the results of their papers. An extensive explanation of the baseline models can be found in Section \uppercase\expandafter{\romannumeral2}.

\begin{table*}[t]
  \centering
  \caption{Performace (in percentage) for entity prediction on YAGO, WIKI, ICEWS18 and GDELT. \\ The best results are bolded, and the second-best results are underlined. }
  \label{tab:results}
  \setlength{\tabcolsep}{0.5mm}{
  \resizebox{1.0\linewidth}{!}{
  \begin{tabular}{c|cccc|cccc|cccc|cccc}
    \toprule
    \multirow{2}{*}{Model} & \multicolumn{4}{c|}{YAGO} & \multicolumn{4}{c|}{WIKI} & \multicolumn{4}{c|}{ICEWS18} & \multicolumn{4}{c}{GDELT} \\
    & MRR & Hits@1  & Hits@3 & Hits@10 
    & MRR & Hits@1  & Hits@3 & Hits@10 
    & MRR & Hits@1  & Hits@3 & Hits@10 
    & MRR & Hits@1  & Hits@3 & Hits@10\\
    \midrule
    TransE & 48.97 & 46.23 & 62.45 & 66.05 
            & 46.68 & 36.19 & 49.71 & 51.71 
            & 17.56 & 2.48 & 26.95 & 43.87 
            & 16.05 & 0.00 & 26.10 & 42.29 \\

    DistMult & 59.47 & 52.97 & 60.91 & 65.26 
            & 46.12 & 37.24 & 49.81 & 51.38 
            & 22.16 & 12.13 & 26.00 & 42.18 
            & 18.71 & 11.59 & 20.05 & 32.55 \\

    ComplEX & 61.29 & 54.88 & 62.28 & 66.82 
            & 47.84 & 38.15 & 50.08 & 51.39 
            & 30.09 & 21.88 & 34.15 & 45.96 
            & 22.77 & 15.77 & 24.05 & 36.33 \\

    ConvE & 62.32 & 56.19 & 63.97 & 65.60
            & 47.57 & 38.76 & 50.10 & 51.53 
            & 36.67 & 28.51 & 39.80 & 50.69 
            & 35.99 & 27.05 & 39.32 & 49.44 \\

    RotatE & 65.09 & 57.13 & 65.67 & 66.16
            & 50.67 & 40.88 & 50.71 & 50.88 
            & 23.10 & 14.33 & 27.61 & 38.72 
            & 22.33 & 16.68 & 23.89 & 32.29 \\

    RGCN & 41.30 & 32.56 & 44.44 & 52.68
            & 37.57 & 28.51 & 39.66 & 41.90 
            & 23.19 & 16.36 & 25.34 & 36.48 
            & 23.31 & 17.24 & 24.96 & 34.36 \\

    CompGCN & 41.42 & 32.63 & 44.59 & 52.81
            & 37.64 & 28.33 & 39.87 & 42.03 
            & 23.31 & 16.52 & 25.37 & 36.61 
            & 23.46 & 16.65 & 25.54 & 34.58 \\
    \midrule

    TTransE & 32.57 & 27.94 & 43.39 & 53.37
            & 31.74 & 22.57 & 36.25 & 43.45 
            & 8.36 & 1.94 & 8.71 & 21.93 
            & 5.52 & 0.47 & 5.01 & 15.27 \\

    HyTE & 23.16 & 12.85 & 45.74 & 51.94
            & 43.02 & 34.29 & 45.12 & 49.49 
            & 7.31 & 3.10 & 7.50 & 14.95 
            & 6.37 & 0.00 & 6.72 & 18.63 \\

    TA-DistMult & 61.72 & 52.98 & 63.32 & 65.19
            & 48.09 & 38.71 & 49.51 & 51.70 
            & 28.53 & 20.30 & 31.57 & 44.96 
            & 29.35 & 22.11 & 31.56 & 41.39 \\

    DySAT & 43.43 & 31.87 & 43.67 & 46.49
            & 31.82 & 22.07 & 26.59 & 35.59 
            & 19.95 & 14.42 & 23.67 & 26.67 
            & 23.34 & 14.96 & 22.57 & 27.83 \\

    TeMP & 62.25 & 55.39 & 64.63 & 66.12
            & 49.61 & 46.96 & 50.24 & 51.81 
            & 40.48 & 33.97 & 42.63 & 52.38 
            & 37.56 & 29.82 & 40.15 & 48.60 \\

    \midrule
    RE-NET & 65.16 & 63.29 & 65.63 & 68.08
            & 51.97 & 48.01 & 52.07 & 53.91 
            & 42.93 & 36.19 & \underline{54.47} & 55.80 
            & 40.12 & 32.43 & 43.43 & 53.80 \\ 

    HIP & 67.55 & 66.32 & 68.49 & 70.37
            & 54.71 & 53.82 & 54.73 & 56.46 
            & 48.37 & 43.51 & 51.31 & 58.49 
            & 52.76 & 46.35 & 55.31 & 61.87 \\ 

    TiRGN & 87.95 & 84.34 & 91.37 & \underline{92.92}
            & 81.65 & 77.77 & 85.12 & 87.08 
            & 33.66 & 23.19 & 37.99 & 54.22 
            & 21.67 & 13.63 & 23.27 & 37.60 \\ 

    RE-GCN & 83.27 & 80.02 & 84.94 & 89.00
            & 81.07 & 78.84 & 82.36 & 84.95 
            & 45.67 & 37.62 & 49.19 & \underline{61.18} 
            & 39.72 & 31.93 & 43.14 & 53.46 \\ 

    TITer & 90.48 & \underline{90.25} & 90.46 & 90.81
            & 74.89 & 74.05 & 74.71 & 76.57 
            & 37.00 & 31.14 & 39.05 & 47.96 
            & - & - & - & - \\ 
    \midrule
    GyGNet & 66.58 & 64.26 & 67.98 & 70.16
            & 52.60 & 50.48 & 53.26 & 55.82 
            & 47.83 & 42.02 & 50.71 & 57.72 
            & 51.06 & 44.66 & 54.74 & 61.32 \\ 

    xERTE & 88.75 & 87.88 & 89.30 & 90.38
            & 77.47 & 76.01 & 78.79 & 79.54 
            & 36.47 & 29.60 & 40.26 & 50.41 
            & - & - & - & - \\ 
    
    CENET & 84.13 & 84.03 & 84.23 & -
            & 68.39 & 68.33 & 68.36 & - 
            & 51.06 & \underline{47.10} & 51.92 & 58.82
            & 58.48 & \underline{55.99} & 58.63 & 62.96 \\ 

    CEN & 85.84 & 83.55 & 87.11 & 90.02
            & 83.11 & 81.20 & 84.15 & 86.46 
            & 45.09 & 37.85 & 47.92 & 59.12
            & 43.54 & 36.51 & 46.13 & 56.88 \\ 

    DA-Net & \underline{91.59} & 90.07 & {\bf 92.94} & {\bf 93.43}
            & \underline{84.13} & \underline{81.66} & \underline{86.46} & {\bf 87.37} 
            & {\bf 51.92} & 45.55 & {\bf 55.70} & {\bf 62.62}
            & \underline{58.47} & 51.89 & \underline{62.32} & {\bf 69.82} \\
    \midrule
    AMCEN & {\bf 92.76} & {\bf 92.73} & \underline{ 92.77} & 92.78
            & {\bf 87.09} & {\bf 87.07} & {\bf 87.11} & \underline{87.12} 
            & \underline{51.65} & {\bf 50.45} & 51.21 & 53.87
            & {\bf 65.23} & {\bf 64.93} & {\bf 64.99} & \underline{65.32} \\

    \bottomrule
  \end{tabular}
  }}

\end{table*}

\subsection{Training Settings and Evaluation Metrics}
The effectiveness of AMCEN is evaluated on the mean metrics of the two queries $(s,r,?,t)$ and $(?,r,o,t)$, including the mean reciprocal rank (MRR) and the proportion of correct predictions ranked within top 1/3/10 (Hits@1/3/10). We adjust the hyperparameters by evaluating MRR performance on each validation set. We use Adam optimizer with a learning rate of 0.001 and a weight decay of 0.00,001 to minimize two-stage losses. The first-stage training epoch with the total loss $\mathcal{L}$ is limited to 30 and the second-stage epoch for a binary classifier is limited to 20. The embedding dimension $d$ is set to 200 and the batch size $n$ is set to 1024. Regarding the structural encoder, we configure the number $L$ of CompGCN updating layers as 2, and the dropout rate for each layer as 0.3. For the temporal encoder, we set the time window size $\tau$ to 4, the number of attention heads $h$ to 10, and the trade-off parameter $\beta$ to 0.2. In terms of training strategy, the hyperparameter $\lambda$ between losses  is assigned to 0.6. Finally, our AMCEN model is coded by using PyTorch and trained on a GPU Tesla V100.

\subsection{Results and Analysis}
Based on the model configuration in the \emph{Training Settings and Evaluation Metrics} subsection, we compare the results of our AMCEN model with that of static and temporal baselines for entity prediction on YAGO, WIKI, ICEWS18 and GDELT, as shown in Table \ref{tab:results}. Our model achieves superior or comparable to state-of-the-art performance on most metrics for entity prediction. Static KGR models perform worse than most of temporal KGR models as they remove timestamp information from original datasets. Temporal KGR models such as TTransE, HyTE, TA-DistMult, DySAT and TeMP, are originally designed to complete missing parts within the known time range, so they exhibits inferior performance compared to extrapolation models. Therefore, as an extrapolation model specifically designed for TKGs, it is inevitable that AMCEN's performance surpasses that of the above models, which is consistent with experimental results. 

Although TITer and xERTE provide some level of interpretability to temporal KGR, their high computational complexity limits their ability to handle large-scale datasets such as GDELT. Compared to them, AMCEN not only boasts enhanced reasoning capabilities but also demonstrates lower complexity when dealing with large-scale data. The nice performance of GyGNet highlights the importance of historical repetitiveness. Likewise, the exceptional results of RE-NET underscore the significance of 1-hop neighbor patterns. Re-GCN, HIP and TiRGN outperform GyGNet and RE-NET because they are able to encode fine-grained temporal evolution and comprehensive structural dependencies. AMCEN covers the advantages of feature extraction in GyGNet, RE-NET, Re-GCN, HIP and TiRGN, thus showcasing superior performance. This is also the reason why AMCEN performs better than CENET. 

\begin{table*}[t]
  \centering
  \caption{Results (in percentage) of ablation studies for entity prediction on  WIKI and ICEWS18. }
  \label{tab:ablation}
  \setlength{\tabcolsep}{0.5mm}{
  \resizebox{0.65\linewidth}{!}{
  \begin{tabular}{c|cccc|cccc}
    \toprule
    \multirow{2}{*}{Model} & \multicolumn{4}{c|}{WIKI} & \multicolumn{4}{c}{ICEWS18} \\
    & MRR & Hits@1  & Hits@3 & Hits@10 
    & MRR & Hits@1  & Hits@3 & Hits@10 \\
    \midrule
    AMCEN-w/o-AM & 79.703 & 71.161 & 82.992 & 86.104 
            & 4.739 & 1.057 & 3.361 & 10.900 \\
    AMCEN-Nonhis & 76.501 & 71.161 & 81.009 & 84.848 
            & 4.248 & 1.399 & 3.511 & 9.143\\
    AMCEN-His & 86.594 & 86.580 & 86.598 & 86.608
            & 51.500 & 50.293 & 51.061 & 53.730 \\
    AMCEN-w/o-PM & 87.047 & 87.037 & 87.040 & 87.046
            & 50.449 & 50.430 & 50.433 & 50.447 \\
    AMCEN & {\bf 87.095} & {\bf 87.069} & {\bf 87.107} & {\bf 87.121}
            & {\bf 51.650} & {\bf 50.445} & {\bf 51.209} & {\bf 53.866} \\
    \midrule
    AMCEN-GT-PM & 87.095 & 87.069 & 87.107 & 87.121
            & 50.763 & 50.553 & 51.321 & 53.990 \\
    \bottomrule
  \end{tabular}
  }}

\end{table*}

DA-Net and CEN achieve the top two state-of-the-art results due to their consideration of temporal variability. However, AMCEN achieves improvements of 2.66\%, 5.41\%, 3.35\%, 8.94\% in Hits@1 over the best baseline on YAGO, WIKI, ICEWS18 and GDELT, respectively. We believe that the use of attention mask vectors can improve the accuracy of predicting new events, leading to these advantages. There are fewer improved metrics on ICEWS18, only in Hits@1. We speculate that this could be attributed to the dataset's high proportion of new events and limited data volume, which may have prevented a thorough exploration of potential factors influencing these new events. Conversely, compared to other datasets, 
there are a significantly greater magnitude and multiple metrics of performance improvement on GDELT. The main reason may be that GDELT has a relatively high proportion of new events, but its ample amount of data ensures thorough exploration and analysis. For our AMCEN model, the reason why the evaluation metrics on WIKI are worse than those on YAGO may be that the distribution of new and recurring events is inconsistent between the training and testing sets on WIKI. Additionally, AMCEN performs slightly worse in Hits@10 than the best baseline. The reason for this poor situation may be that the binary classifier of the first reasoning stage narrows the exploration scope of the model.

\subsection{Ablation Study}

We conducted ablation experiments on WIKI and ICEWS18 to investigate the effects of removing specific model components from AMCEN. It helps us gain a better understanding of the relative importance of each component in the overall performance of the model. The results of ablation studies are shown in Table \ref{tab:ablation}. 

We use the ground truths available in the test set to create a AMCEN-GT-PM, while maintaining all other variables at their original values. To analyze their influence effectively, we remove the predictive mask vectors during the inference stage to obtain AMCEN-w/o-PM. Our AMCEN model performs equally well as AMCEN-GT-PM on WIKI, but slightly worse than AMCEN-GT-PM on ICEWS18. This may be attributed to the fact that ICEWS18 dataset has the largest number of entity and relation types, leading to poorer performance of the binary classifier. AMCEN outperforms AMCEN-w/o-PM on both ICEWS18 and WIKI, which highlights the significance of predictive mask vectors.

To access the effectiveness of historical/non-historical attention mask vectors, we design three experiments: AMCEN-w/o-AM, AMCEN-His and AMCEN-Nonhis. Regarding AMCEN-w/o-AM, we remove attention mask vectors and treat all entities with equal importance. AMCEN-His only retain a historical attention mask vector, while AMCEN-Nonhis solely rely on a non-historical attention mask vector. The results of AMCEN-w/o-AM significantly decrease compared to AMCEN and even approaches zero on ICEWS18. Specifically, the absence of attention mask vectors results in a decline of approximately 15\% in Hits@1 on WIKI and 49\% on ICEWS18. It is obvious that attention mask vectors are necessary for achieving a satisfactory prediction performance. AMCEN-His outperforms AMCEN-Nonhis, possibly due to a higher prevalence of repeated events and more readily discernible patterns in the data. Additionally, in contrast to AMCEN-w/o-AM, the inferior performance of AMCEN-Nonhis can be attributed to its narrow attention on non-historical entities, which results in disregarding a substantial amount of information and ultimately leads to disastrous inference. Therefore, the combination of the historical attention mask vector and the non-historical attention mask vector can achieve excellent predictive performance, as they are complementary to each other.

\begin{figure*}[htbp]
\centering
\subfigure[Parameter $\tau$ on WIKI]{
\begin{minipage}[t]{0.23\linewidth}
\centering
\includegraphics[width=1.7in]{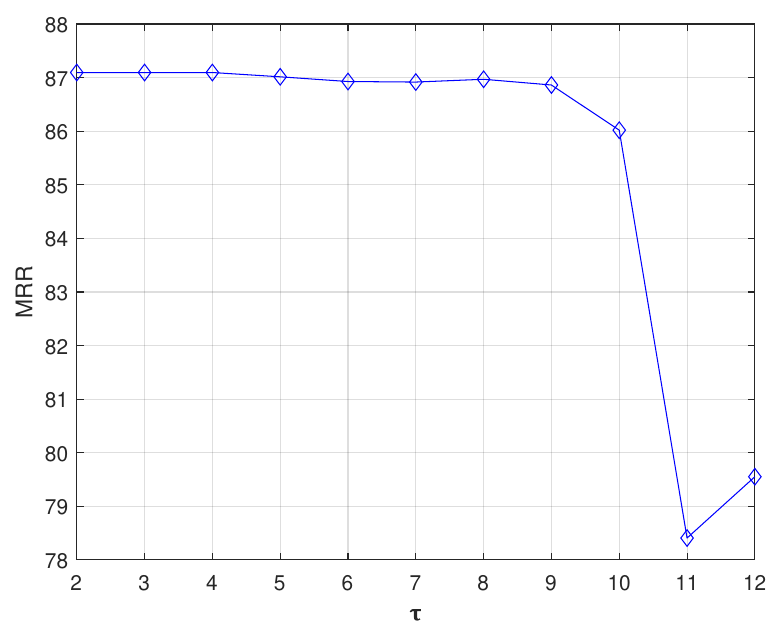}
\end{minipage}%
}%
\subfigure[Parameter $\tau$ on ICEWS18]{
\begin{minipage}[t]{0.23\linewidth}
\centering
\includegraphics[width=1.7in]{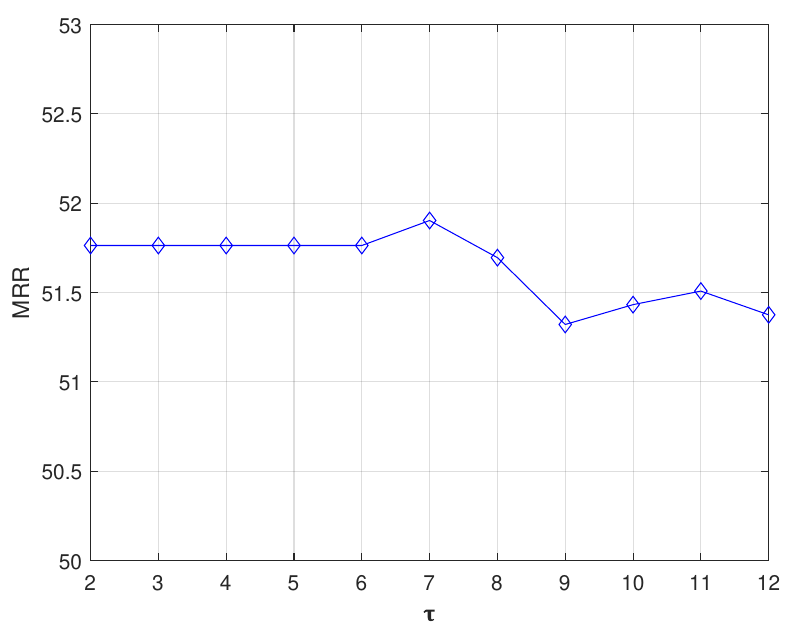}
\end{minipage}%
}%
\subfigure[Parameter $h$ on WIKI]{
\begin{minipage}[t]{0.23\linewidth}
\centering
\includegraphics[width=1.7in]{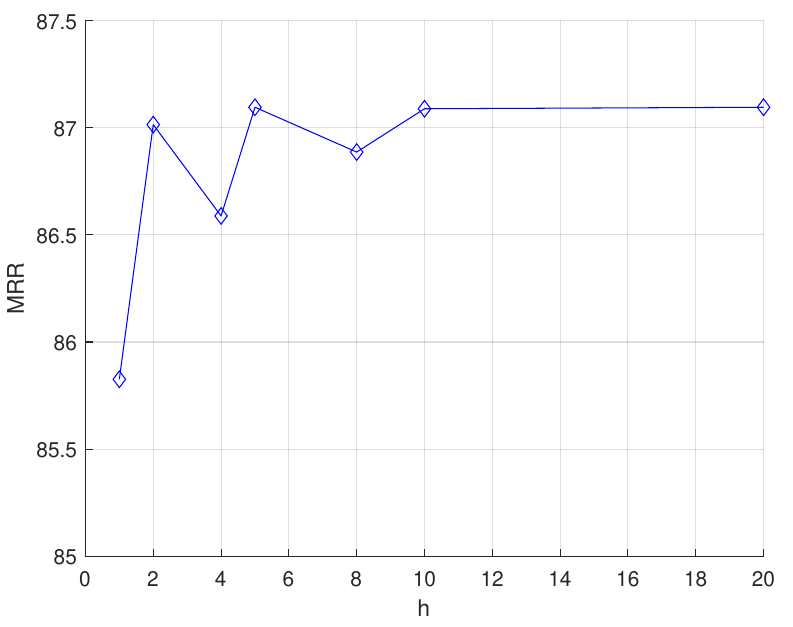}
\end{minipage}
}%
\subfigure[Parameter $h$ on ICEWS18]{
\begin{minipage}[t]{0.23\linewidth}
\centering
\includegraphics[width=1.7in]{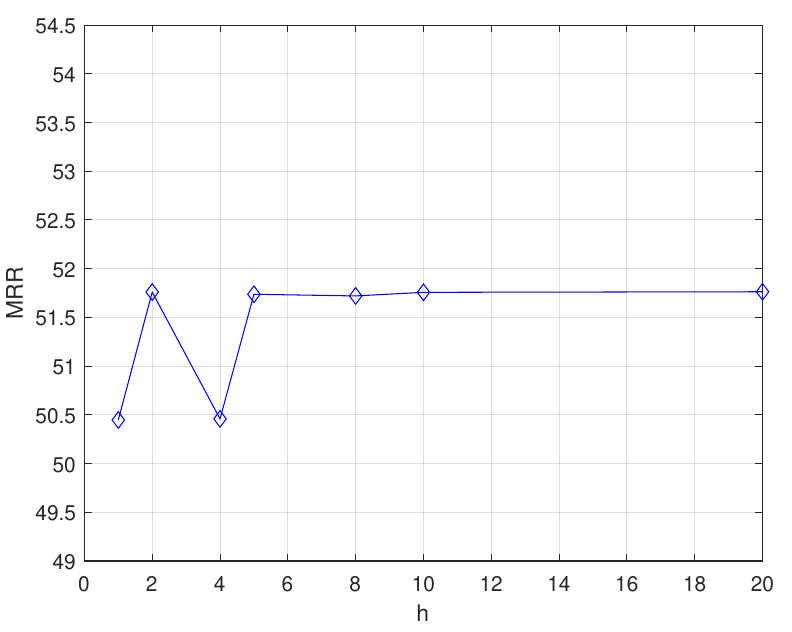}
\end{minipage}
}%

\subfigure[Parameter $\beta$ on WIKI]{
\begin{minipage}[t]{0.23\linewidth}
\centering
\includegraphics[width=1.7in]{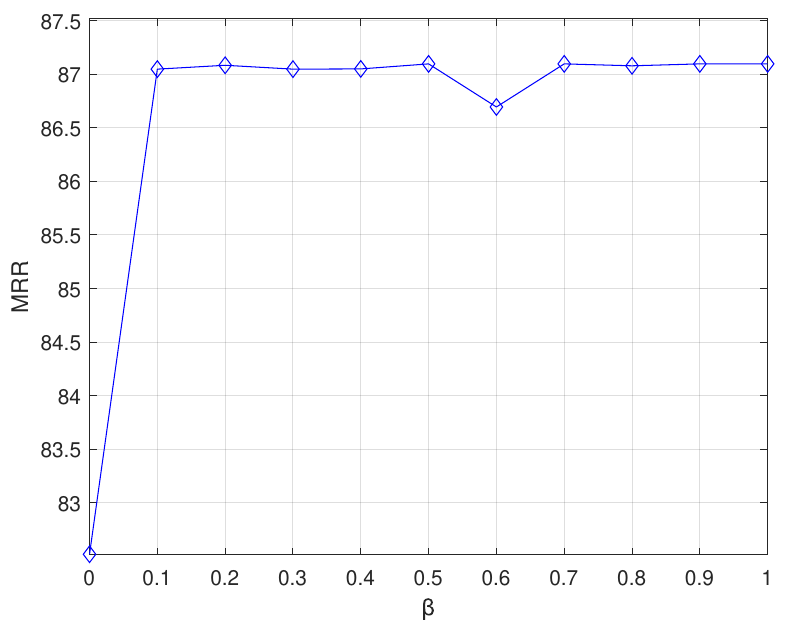}
\end{minipage}%
}%
\subfigure[Parameter $\beta$ on ICEWS18]{
\begin{minipage}[t]{0.23\linewidth}
\centering
\includegraphics[width=1.7in]{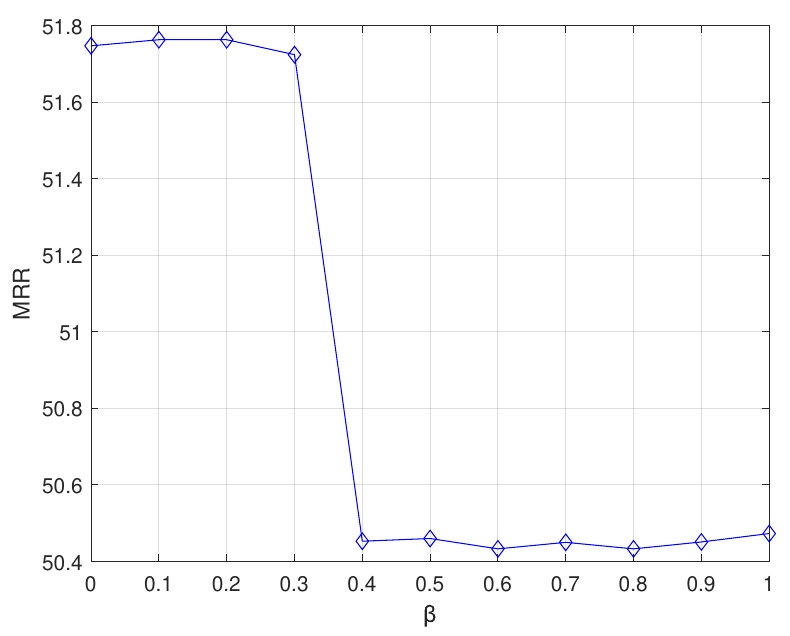}
\end{minipage}%
}%
\subfigure[Parameter $\lambda$ on WIKI]{
\begin{minipage}[t]{0.23\linewidth}
\centering
\includegraphics[width=1.7in]{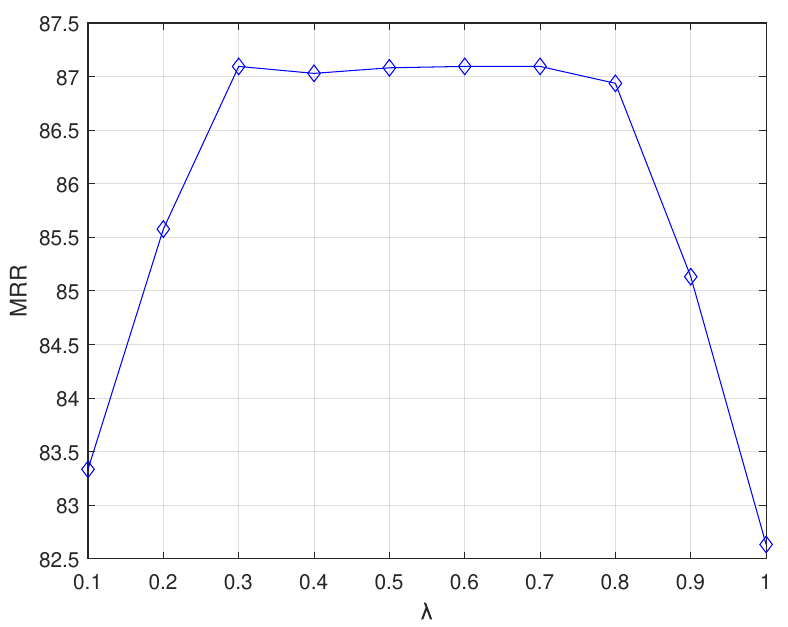}
\end{minipage}
}%
\subfigure[Parameter $\lambda$ on ICEWS18]{
\begin{minipage}[t]{0.23\linewidth}
\centering
\includegraphics[width=1.7in]{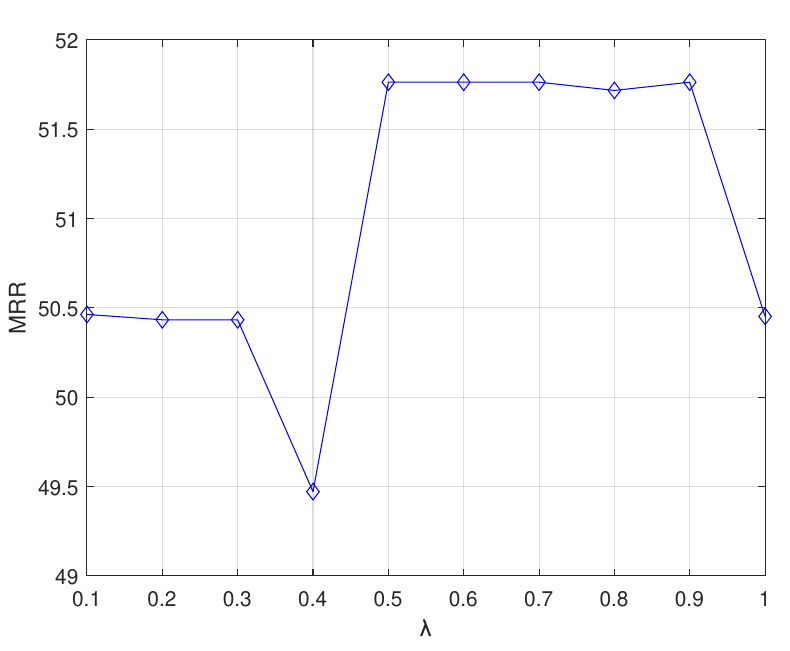}
\end{minipage}
}%
\centering
\caption{Sensitivity analysis results of hyperparameters in MRR}
\label{hyperparameter}
\end{figure*}

\subsection{Hyperparameter Analysis}

We explore the influence of the hyperparameters $\tau$, $h$, $\beta$ and $\lambda$ for the MRR performance of the model using the WIKI and ICEWS18 datasets, as shown in Fig. \ref{hyperparameter}. 

The hyperparameter $\tau$ controls the capturing range of adjacent temporal sequences for the local temporal encoder. As shown in Fig.\ref{hyperparameter} (a) and (b), if the value of $\tau$ is set too large, the MRR results of our model may decrease as a result of an excessive amount of irrelevant information being introduced. The performance is better when the value of $\tau$ is from 2 to 7.

The hyperparameter $h$ indicates the number of perspectives from which local temporal patterns are learned. Fig.\ref{hyperparameter} (c) and (d) demonstrate that the model shows an overall upward trend in MRR as $h$ increases. The change of $h$ from 10 to 20 does not result in a significant performance improvement while a noticeable increase in the number of parameters. Therefore, the optimal choice for $h$ is 10. 

The hyperparameter $\beta$ aims at adjusting the impact of information from the timestamp $t-1$ on the event prediction at the timestamp $t$. Due to the distinct characteristic of event-based TKGs and fact-based KGs, Fig.\ref{hyperparameter} (e) and (f) show different trends for MRR curves. In WIKI, a fact (also called an event) often lasts a long time, but not so in ICEWS18. Therefore, the removal of the timestamp $t-1$ ($\beta$=0) has a significant effect on the MRR performance of the model on WIKI, while it has no effect on ICEWS18. When $\beta$ is equal to or larger than 0.4, our model exhibits relatively poor MRR results on ICEWS18, which proves that excessive information from the timestamp $t-1$ can even mislead the future prediction. Obviously, considering both two types of datasets together, selecting a $\beta$ value ranging from 0.2 to 0.3 would be more appropriate. 

The objective of setting the hyperparameter $\lambda$ is to strike a balance between the contributions of $\mathcal{L}_{mc}$ and $\mathcal{L}_{bc}$. As shown in Fig. \ref{hyperparameter} (g) and (h), if the value of $\lambda$ is small, our model's MRR performance drops. It proves that less contribution from $\mathcal{L}_{mc}$ can result in inadequate mining of temporal evolution and structural dependencies, ultimately leading to inferior performance. Likewise, the lack of $\mathcal{L}_{bc}$ ($\lambda=1$) results in poor MRR performance.  This is because the removal of adversarial training weakens the binary classifier's ability to classify, which then causes inaccuracies in masking the predicted probability distribution.

\section{Conclusion and Future Work}
In this paper, to tackle the problem of imbalanced event proportions, we propose AMCEN for two-stage reasoning, an attention masking-based contrastive event network with local-global temporal patterns. In this model, historical and non-historical attention mask vectors are created to achieve the separation of learning and mining of potential factors related to recurring and new events.
The integration of CompGCN and self-attention allows the model to explore structural interactions and temporal evolution between entities and relations. It can deal with temporal sparsity and ensure diminishing historical effects. Additionally, we incorporate local-global patterns into the contrastive learning framework to refine the scope of prediction. Experiment results on four datasets show the advantages and effectiveness of AMCEN for entity prediction. And ablation experiments show attention mask vectors and predictive mask vectors play a positive and crucial role in Temporal KGR. 

However, AMCEN also has some limitations: 1) Its unexplainable reasoning process leads to a lack of persuasiveness in specialized fields like medicine and finance; 2) Its understanding of relationship correlations and temporal changes is not strong enough, resulting in decreased accuracy of predictions on limited datasets like ICEWS18. Large-scale language models (LLM)\cite{lu2023chatgpt} have garnered increasing attention thanks to their remarkable language comprehension and emerging logical reasoning abilities. Chain-of-thought Prompting \cite{wei2022chain} breaks down a problem-solving process into a series of logical reasoning steps, thereby enhancing the capacity of large language models to effectively handle intricate forms of reasoning. Additionally, in contrast to feature engineering, scenarios engineering \cite{li2022features,Yang2023DeFACT,li2022novel} seamlessly combines intelligence and index (I\&I), calibration and certification (C\&C) and verification and validation (V\&V). The approach ensures the visibility of individuals into the operation of a machine, and the controllability of the learning features. Therefore, in the future, we will explore the integration of LLM with scenarios engineering to achieve trustworthy high-accuracy KGR while ensuring reasoning speed.

\section*{Acknowledgments}
This work was sponsored by ZhejiangLab Open Research Proiect (No.K2022KG0AB02).

% {\appendix[Proof of the Zonklar Equations]
% Use $\backslash${\tt{appendix}} if you have a single appendix:
% Do not use $\backslash${\tt{section}} anymore after $\backslash${\tt{appendix}}, only $\backslash${\tt{section*}}.
% If you have multiple appendixes use $\backslash${\tt{appendices}} then use $\backslash${\tt{section}} to start each appendix.
% You must declare a $\backslash${\tt{section}} before using any $\backslash${\tt{subsection}} or using $\backslash${\tt{label}} ($\backslash${\tt{appendices}} by itself
%  starts a section numbered zero.)}

%{\appendices
%\section*{Proof of the First Zonklar Equation}
%Appendix one text goes here.
% You can choose not to have a title for an appendix if you want by leaving the argument blank
%\section*{Proof of the Second Zonklar Equation}
%Appendix two text goes here.}

\bibliographystyle{IEEEtran}      %IEEEtran为给定模板格式定义文件名
\bibliography{main.bib}

\vfill

\end{document}